\documentclass[10pt,journal,compsoc]{IEEEtran}
\ifCLASSOPTIONcompsoc
  \usepackage[nocompress]{cite}
\else
  \usepackage{cite}
\fi

\usepackage{mathptmx}
\usepackage{graphicx}
\usepackage{times}
\usepackage{caption}
\usepackage{morefloats}
\usepackage[tight,normalsize,sf,SF]{subfigure}

\usepackage{amsmath,amscd}
\usepackage{amsfonts}
\usepackage[ruled,vlined]{algorithm2e}
\usepackage{algorithmic}
\usepackage{epsfig}
\usepackage{enumerate}
\usepackage{multirow}
\usepackage{array}
\usepackage{diagrams}
\usepackage{diagbox}

\newtheorem{property}{Property}

\usepackage{color,soul} 
\usepackage{makecell}

\makeatletter
 \def\SOUL@hlpreamble{%
 \setul{}{2.4ex}
 \let\SOUL@stcolor\SOUL@hlcolor
 \SOUL@stpreamble
 }
\makeatother

\begin{document}
%

\title{FeatureLego: Volume Exploration Using Exhaustive Clustering of Super-Voxels}

%
%
%
%

\author{Shreeraj~Jadhav,~Saad~Nadeem,~and~Arie~Kaufman,~\IEEEmembership{Fellow,~IEEE}
\IEEEcompsocitemizethanks{\IEEEcompsocthanksitem Shreeraj Jadhav and Arie Kaufman are with the Department of Computer Science, Stony Brook University, Stony Brook, NY 11794-2424. Saad Nadeem is with the Department of Medical Physics, Memorial Sloan Kettering Cancer Center, NY 10065.
\protect\\
E-mail: \{sdjadhav, ari\}@cs.stonybrook.edu; nadeems@mskcc.org
}
\thanks{}}
\IEEEtitleabstractindextext{%
\begin{abstract}
We present a volume exploration framework, FeatureLego, that uses a novel voxel clustering approach for efficient selection of semantic features.
We partition the input volume into a set of compact super-voxels that represent the finest selection granularity.
We then perform an exhaustive clustering of these super-voxels using a graph-based clustering method.
Unlike the prevalent brute-force parameter sampling approaches, we propose an efficient algorithm to perform this exhaustive clustering.
By computing an exhaustive set of clusters, we aim to capture as many boundaries as possible and ensure that the user has sufficient options for 
efficiently selecting semantically relevant features.
Furthermore, we merge all the computed clusters into a single tree of meta-clusters that can be used for hierarchical exploration.
We implement an intuitive user-interface to interactively explore volumes using our clustering approach.
Finally, we show the effectiveness of our framework on multiple real-world datasets of different modalities.
\end{abstract}

}

\maketitle

\IEEEdisplaynontitleabstractindextext

%
\IEEEpeerreviewmaketitle

\section{Introduction}
    Approaches that use clustering of voxels over histograms and volumes have been established as effective tools for interactive volume exploration. 
These clusters can classify voxels into different materials, and can capture important regions and boundaries of semantic features.
Such clustering of voxels can be used to apply optical properties for rendering and to visualize individual features during the exploration process.
Previous approaches often compute a single flat or hierarchical clustering which may not always separate semantic features. 
A user is then required to either modify the cluster boundaries interactively, or re-execute the clustering algorithm with different parameters.
To overcome this limitation, multiple clusterings can be generated by sampling the input parameter space of the clustering algorithm. 
However, parameter sampling is not exhaustive~(e.g., \figurename~\ref{fig:range_tracking}) 
and hence, may require significant trial and error to obtain the correct samples. 
Additionally, dense sampling can produce duplicate clusterings and increase the pre-processing computation.

In this paper, we propose an efficient algorithm to exhaustively cluster super-voxels into regions of different sizes.
These regions can be used as building blocks for efficient selection and visualization of features, hence the name FeatureLego.
By computing clusters exhaustively, we are more likely to capture all the important boundaries and regions 
as compared to previous approaches that use a single flat or hierarchical clustering.
The advantages of using exhaustive clustering for volume exploration are as follows.
First, it provides more flexibility to the user for selecting features-of-interest.
Second, it alleviates the burden of guessing the right clustering parameters. 
Third, it saves time by eliminating the trial-and-error approach of parameter sampling.
Fourth, by providing an exhaustive result-set of clusters, it allows for a comprehensive exploration of data.

In our framework, we first compute smaller compact super-voxels over the input volume using 
a 3D extension (termed 3D SLIC) of the simple linear iterative clustering (SLIC)~\cite{Achanta:2012:Slic},
which employs a k-means clustering to efficiently generate superpixels. 
We then perform an exhaustive clustering of these super-voxels using a 3D extension of Felzenszwalb and Huttenlocher (FH) method~\cite{Felzenszwalb:2004:GBH}.
~The FH method is a graph-based clustering algorithm that uses a greedy strategy to progressively aggregate graph-nodes into clusters. 
It uses a single input parameter $k$, that implicitly controls the size of generated clusters.
The FH method is an established algorithm for computing super-pixels in 2D images and is known to capture boundaries and regions well.
We propose an algorithm to efficiently compute all unique clusters that can be computed using the FH method for the parameter space $k \in \lbrack 0,\infty )$ in 3D data.
We achieve this without brute force parameter sampling by tracking maximal contiguous intervals of $k$.
For managing this large set of regions during user interaction, 
we group together regions with high degree of overlap and construct a hierarchy of these \emph{meta-clusters} using containment.
This tree of meta-clusters can be used for efficient exploration, and for performing filtering and search queries during the exploration process.

To the best of our knowledge, we are the first to introduce exhaustive clustering for volume exploration.
Based on this approach, we describe a visual exploration framework for complex volumetric data of different modalities. 
We present an efficient parallel implementation of our algorithm and implement an intuitive user-interface for exploration.
The contributions of the paper are summarized as follows:
\vspace{-2mm}
\begin{itemize}
    \item An efficient exhaustive algorithm to compute all possible outputs that can be generated using the 3D extension of the FH clustering without brute-force parameter sampling.
    \item A novel technique to merge multiple segmentations of a volume into a hierarchy of overlapping regions. 
    \item An efficient parallel implementation of the proposed exploration pipeline and a user-interface for volume exploration.
    \item An evaluation of our exploration framework on complex real-world volumes from different modalities.
\end{itemize}
\begin{figure*}[!ht]
\centering
\includegraphics[width=\textwidth]{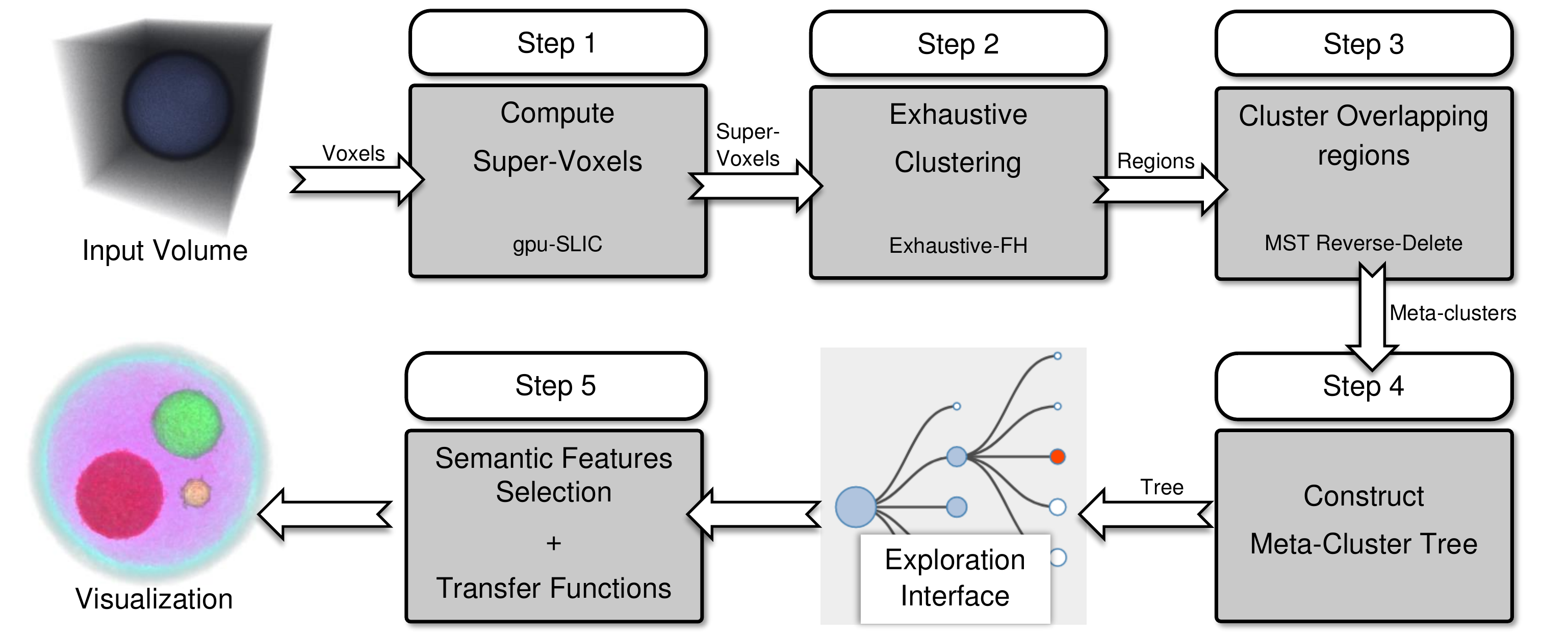}
\caption{FeatureLego Pipeline: The input volume is first partitioned into compact super-voxels using 3D SLIC.
         Super-voxels are clustered exhaustively using our proposed exhaustive-FH algorithm to construct selectable regions.
         Overlapping regions are grouped into meta-clusters using Minimal-Spanning-Tree (MST) based reverse-delete clustering algorithm. 
         A hierarchy of these meta-clusters is constructed
         using our tree construction algorithm (described in section~\ref{sec:tree}).
         The user generates desired visualizations by selecting single or multiple nodes from the meta-cluster tree as features-of-interest.
         Optical properties (e.g., transfer function) are applied locally to each selected feature.}
         \label{fig:framework}\vspace{-3mm}
\end{figure*}

\section{Related Work}
        General techniques for volume exploration and visualization are well studied. 
    However, separating semantic features in complex datasets is challenging and remains an important open problem.
    
    \textbf{Histogram clustering.} 
    A common approach to separating features and material classes is to cluster voxels in 2D and higher dimensional histograms~\cite{Tzeng:2004:CVI}. 
    Sereda et al. \cite{Sereda:2006:ATF} have applied hierarchical clustering to extract features of different sizes over the 2D LH histogram \cite{Sereda:2006:LHHistograms}. 
    Roettger et al. \cite{Roettger:2005:STF} have used spatial information in the 2D histograms to isolate spatially disconnected features (clusters) 
    that might otherwise have similar voxel properties. 
    Non-parametric clustering of 2D histograms has also been proposed \cite{Maciejewski:2009:SFS}, which allowed the user to 
    explore feature boundaries by controlling the sizes of the clusters interactively. 
    Wang et al.~\cite{Wang:2011:GMMs:TVCG} have used the Gaussian Mixture Model to separate feature voxels over 2D histograms.
    Ip et al.~\cite{Ip:2012:hierarchical} have applied the normalized-cut technique to recursively segment 2D intensity-gradient histograms into a binary hierarchy. 
    Modified dendrograms~\cite{Wang:2012:MultiDimTFDesign} were used to facilitate user interaction over hierarchical clustering of higher-dimensional feature spaces.
    Ponciano et al.~\cite{Ponciano:2016:GraphBasedExplore} and Wang et al.~\cite{Wang:CGF2012:ValleyCellTF} have applied topological segmentation and simplification 
    over 2D histograms to aggregate regions into features. 
    However, clustering over feature spaces and histograms is often limited because they do not consider spatial connectivity information during clustering.
    Histograms have limited precision and are known to have limited ability in differentiating features~\cite{Carr:2006:isostats}.
    
    \textbf{Volume segmentation.}
    Some approaches directly cluster voxels or segment features over the spatial domain to facilitate feature selection. 
    These techniques have an advantage  as they consider spatial connectivity when separating features.
    Xiang et al.~\cite{Xiang:2011:SkeletonCuts} have used a combination of segmentation and local transfer functions to visualize complex medical datasets.
    The segmentation provides better localization of transfer functions and better feature separation.
    Topology-based approaches~\cite{weber2007topology,Guo:2011:WYSIWYG,Woo:2012:Feature} use iso-surfaces and contour trees for direct selection of features in a volume. 
    However, selection can be hard to control and tedious in noisy data or in the case of complex features.
    Shen et al.~\cite{Shen:2015:ModelDriven} have proposed the use of a pre-labeled semantic model to segment and render features of an input volume.
    Simpler segmentation approaches such as region growing and thresholding~\cite{Huang:Ma:RGVis,Hadwiger:2008:InteractiveExp,Reh:2013:MObjects}
    have also been proposed for feature extraction in volumes.
    
    \textbf{Parameter Sampling.}
    Parameter sampling approaches for visualization and exploration of data focus on searching for appropriate rendering and segmentation parameters. 
    Interfaces such as Design Galleries~\cite{Marks:1997:DGG} and stochastic search techniques~\cite{he1996generation} were developed for sampling and exploration of appropriate rendering parameters.
    Bruckner and M\"oller~\cite{Bruckner:2010:SimParamSpaces} have used parameter space exploration approach to identify simulation parameters for visual effect design.
    Image segmentation and analysis were also performed using parameter sampling~\cite{Pretorius:2011:VisParamIA,Torsney:2011:Tuner,Frohler:2016:GEMSe}.
    A Conceptual Framework for parameter sampling was described by Sedlmair et al.~\cite{Sedlmair:2014:VPSAF}.
    Parameter sampling approaches are by their very nature not exhaustive and hence can miss out on features in the final output. 
    
    \textbf{Graph Representations.}
    For exploring spatial data, features are often organized into a tree or a graph
    to support searching for features-of-interest and to understand relationships between them.
    Hierarchical clustering and segmentation approaches~\cite{Sereda:2006:LHHistograms,Ip:2012:hierarchical} naturally organize features into a tree.
    Similarly, contour trees~\cite{Carr:2003:ComputingContourTrees,Pascucci:2004:ContourTrees} also organize features into a tree based on the morphological behaviour of level-sets.
    Balabanian et al.~\cite{Balabanian:2010:IIV} have used graph-based 
    illustrative visualization for interactive exploration.
    Wang et al.~\cite{Wang:2005:Hierarchical} have utilized a hierarchical interface for controlling and rendering
    level-of-detail in simulation data.
    A graph-based representation of spatial relations between extracted features was presented by Chan et al.~\cite{Chan:2008:RelationAware}.
    Similarly, Gu et al.~\cite{Gu:2011:TransGraph,gu2013itree} have developed a graph representation for visualizing transition relationships in time-varying data.
    A recent survey on graph-based representations in visualization was presented by Wang and Tao~\cite{Wang:2017:Graphs}.
    We present an algorithm to organize any given set of multiple segmentations into an exploration hierarchy.
    The tree is constructed to make exploration efficient and intuitive by grouping overlapping regions together,
    guaranteeing contained regions as descendants, and sorting sibling nodes by size for guided search.


\section{FeatureLego Framework}
    Our framework performs exhaustive clustering of super-voxels to produce multiple selectable regions and constructs an exploration tree for the user to efficiently select and visualize semantic features.
The motivation for performing an exhaustive clustering is to free the user from the burden of guessing the right clustering parameters and to capture as many semantic features as possible in the form of regions. 
This makes the exploration process more time-efficient, since the user does not have to perform a time consuming trial-and-error process to determine the right clustering parameters for capturing desired features. Additionally, a manual search is not exhaustive and consequently less likely to extract all the desired features-of-interest.
We extend clustering techniques from the field of image analysis to volumetric data as these techniques are known to capture perceptually homogeneous regions well.
The extracted regions are then arranged into an exploration hierarchy.
We call the constructed exploration structure a meta-cluster tree, since each node represents a group of regions (cluster of clusters) that have a high degree of overlap.
This meta-cluster tree helps the user to efficiently explore similar regions that are generated through exhaustive clustering.
The user can select individual or multiple regions as a semantic feature (i.e. a feature meaningful to the user's domain) by navigating through the tree.

\begin{figure}
    \centering
    \includegraphics[width=0.95\linewidth]{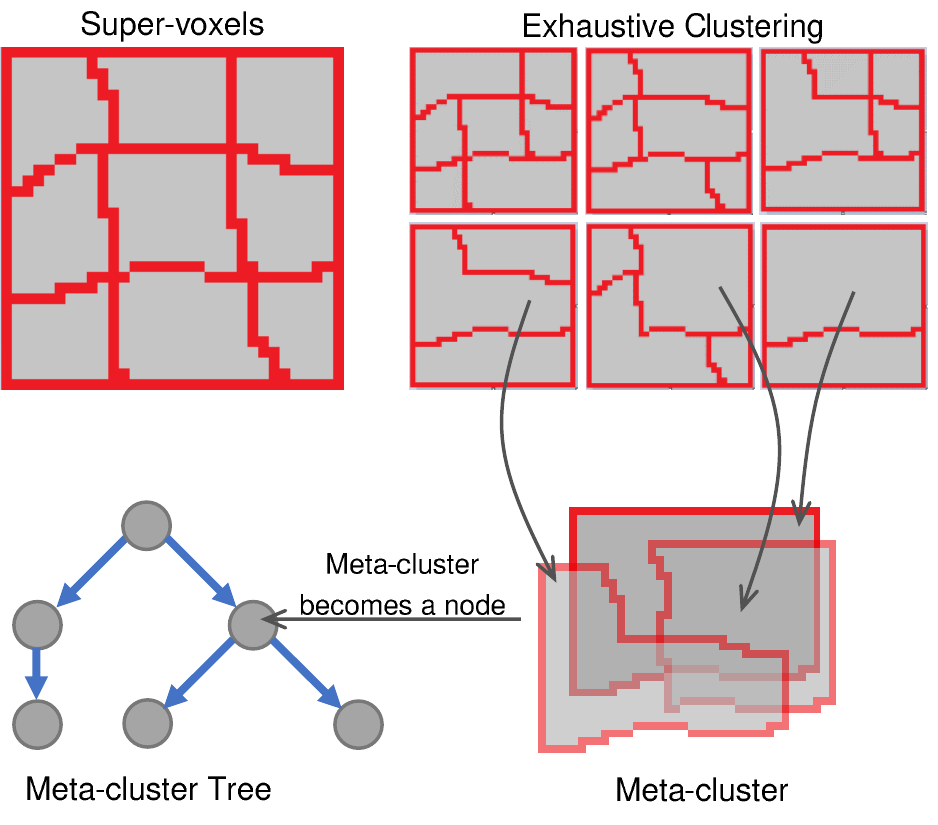}
    \caption{An illustration of different levels of clustering in FeatureLego. First, the voxels of input volume are clustered into super-voxels.
    Second, exhaustive clustering is performed to compute regions as a clustering of super-voxels.
    Finally, regions having significant overlap are grouped together to form meta-clusters, which are then used as nodes of the meta-cluster tree.}
    \label{fig:clustering_illustration}\vspace{-4mm}
\end{figure}

The FeatureLego pipeline consists of five main steps as illustrated in \figurename~\ref{fig:framework}.
Steps 1 and 2 are performed for extracting features, Steps 3 and 4 facilitate exploration; and Step 5 is interactive exploration performed by the user through our implemented interface.
In Step 1, the computation of compact super-voxels is performed using a GPU based implementation of 3D SLIC~\cite{Achanta:2012:Slic} algorithm.
The SLIC algorithm provides explicit control over compactness and size of super-voxels through input parameters.
These super-voxels are the smallest regions a user can select in our framework to visualize semantic features.
The reasons for using SLIC super-voxels as base granularity are twofold. 
First, they provide local compactness to larger regions computed through exhaustive-FH clustering.
Compactness is desirable to reduce fragmentation in computed regions and to facilitate intuitive selection.
Second, they make the exhaustive clustering more efficient and scalable to real-world datasets as the initial adjacency graph constructed by the 
exhaustive-FH method is smaller in size.
In Step 2, an exhaustive clustering of these super-voxels is performed using our exhaustive-FH algorithm. This step merges super-voxels in different combinations to produce regions as prospective features-of-interest.
In Step 3, we use a Minimal Spanning Tree (MST) based clustering algorithm called \emph{reverse-delete} to group together regions that have a high degree of overlap.
We refer to such a group of regions as a meta-cluster.
In Step 4, we construct a meta-cluster tree to facilitate hierarchical exploration by organizing the computed meta-clusters as individual nodes of the tree.
Parent-child relationships between nodes (meta-clusters) of the tree are established through containment. 
A parent node is the smallest meta-cluster that contains the child node.
It is important to note that our pipeline contains exactly three levels of clustering.
The first level clusters voxels of the input volume into super-voxels using SLIC.
The second level clusters super-voxels into regions using exhaustive-FH.
And, the third level clusters regions into meta-clusters using the MST-based reverse-delete algorithm. 
We use this terminology consistently throughout the paper to refer to the different stages of our pipeline.
An illustrative example is shown in \figurename~\ref{fig:clustering_illustration} that demonstrates each step towards constructing the meta-cluster tree.

Additionally, we implement a user-interface to explore and visualize features-of-interest using the meta-cluster tree.
The tree is rendered as a collapsible graph from which nodes and regions can be selected interactively.
Selected regions are added as bookmarks into a list of semantic features. 
Optical properties for each region can be set separately through a pop-up window.
Details of the exhaustive-FH clustering algorithm, the meta-cluster tree construction, and the implemented exploration interface are provided in the following subsections.
An accompanied supplementary video shows the exploration interface in action.


\subsection{Exhaustive-FH Clustering}
    While the FH method was originally described for clustering pixels in 2D images, 
Grundmann et al.~\cite{Grundmann:2010:GBVS} have extended this method for segmenting larger regions in videos.
We extend the FH method to perform exhaustive clustering of super-voxels in 3D volume data.
We give a brief overview of the original technique in the context of 2D images, and describe how we extended it to volume data for exhaustive clustering of super-voxels.
\subsubsection{Overview of the FH Method}  
    The FH method, as originally applied to 2D images, constructs an adjacency graph where each node of the graph is a pixel and each edge connects immediate neighbors in the input image. 
    Edges are assigned weights based on the intensity values of the connected pixels. 
    Low weights indicate high similarity between pixels, while high values indicate low similarity.
    Typically, edge-weights are simply computed as the absolute value of intensity differences between pixels.
    Pixels are then merged iteratively into larger regions based on these edge weights. 
    The algorithm iterates through the list of edges in a non-decreasing order of weights. 
    Adjacent regions are merged if the minimum edge weight that connects them is smaller than the internal variation of edge weights within these regions. 
    For a given region $C$, internal variation $\text{Int}(C)$ is estimated as the smallest edge weight in the minimum spanning tree of that region. 
    Therefore, two adjacent regions $C_1$ and $C_2$ are merged if the following predicate is true:
    \[
     w(C_1, C_2) \leq \text{min} \bigg( \text{Int}(C_1), \text{Int}(C_2) \bigg),
    \]
    where $w(C_1, C_2)$ is the minimum edge weight connecting regions $C_1$ and $C_2$.
    Initially, each region is a single pixel, and hence, does not contain any internal edges. 
    This prohibits pixels from merging as their internal variation is zero. 
    A parameter $\tau(C_i)$ is defined for each region $C_i$ to offset this value and initiate merging, where
    \[
    \tau(C_i) = {\frac{k}{\vert C_i \vert} }
    \]
    and $\vert C_i \vert$ is the size of the region $C_i$. The resultant merging condition is expressed as,
    \begin{equation}\label{eqn:merging_predicate}
     w(C_1, C_2) \leq \text{min} \bigg( \text{Int}(C_1) + \tau(C_1), \text{Int}(C_2) + \tau(C_2) \bigg)
    \end{equation}
    The value of $k$ is provided by the user and it implicitly controls the size of the resulting regions. Values of region size $|C|$ and parameter $\tau$ have to be updated after every merge operation of FH method. While region size is a simple addition after the merge process, $\tau$ of the newly created region is simply the weight of the edge that caused that merge. Both the values can be updated in constant time.
    For 2D images, FH constructs the adjacency graph based on an 8-neighbor configuration. 
    Grundmann et al.~\cite{Grundmann:2010:GBVS} have extended this method for videos and used a 26-neighbor configuration to construct the graph.
\subsubsection{Our Algorithm}
\begin{figure}[!t]
\centering \includegraphics[width=\linewidth]{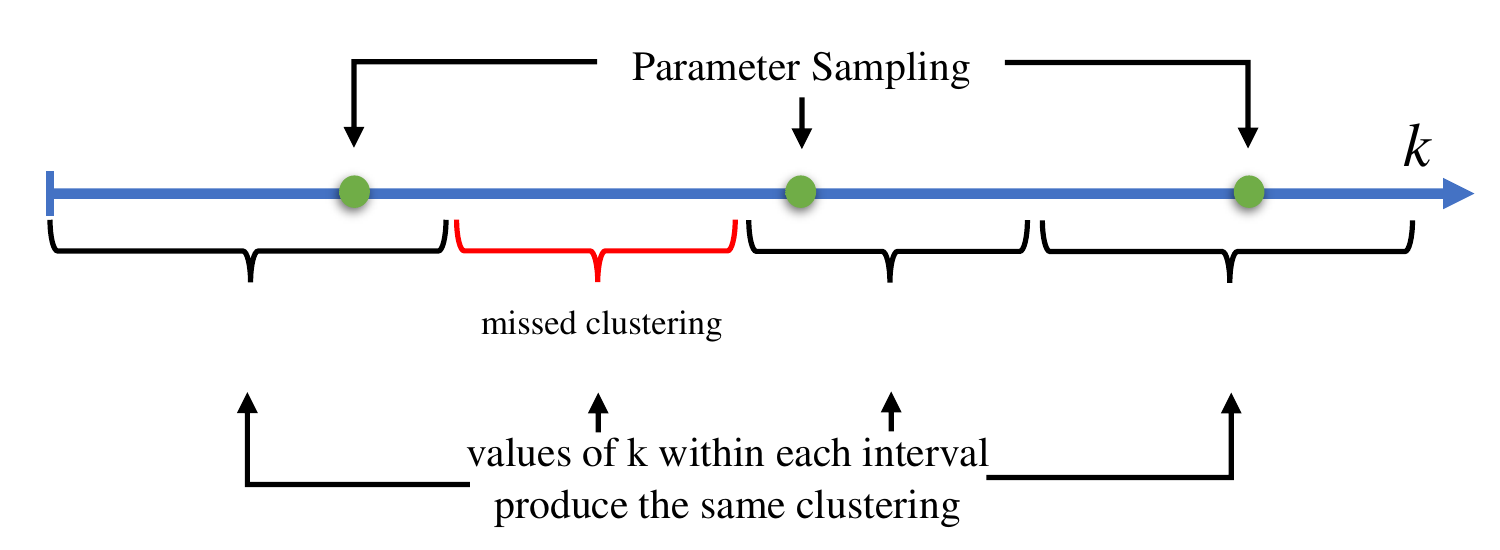}
\caption{
    Input parameter $k$ of the FH method shown as an axis of real numbers.
    Green points: values of $k$ chosen by uniform parameter sampling.
    Brace brackets: Contiguous intervals of $k$ along the axis where any value of $k$ within an interval produces the same clustering. 
    Parameter sampling is not exhaustive and might miss important clustering results.
    Whereas, our approach of interval-tracking guarantees that all unique clusterings will be computed.}
    \vspace{-3mm}
\label{fig:range_tracking}
\end{figure}

    We apply the FH method for clustering pre-computed super-voxels in 3D volumes. This is done by constructing an unstructured graph (not a regular grid) based on adjacency between the super-voxels.
    Since the adjacency graph is not a regular grid, we do not have a fixed neighborhood configuration for every node. The number of edges for each node can vary based on the pre-computed super-voxels.
    Our proposed exhaustive-FH algorithm computes all possible clusterings over this graph that can be generated using the FH method by spanning the entire space of input parameter $k \in \lbrack 0,\infty ) $.
    We achieve this efficiently without brute-force sampling by tracking the maximal contiguous intervals of $k$ using multiple executions of the FH method (\figurename~\ref{fig:range_tracking}).
    All values of $k$ within such an interval produce the same clustering.
    Since parameter $k$ implicitly controls the size of the output regions, for a sufficiently large value of $k$, all nodes of the input graph merge into a single region.
    At this point, our algorithm terminates as no new clustering can be generated using any higher values of $k$. Essentially, our proposed exhaustive-FH algorithm repeatedly executes the FH method while tracking one contiguous interval $[k_s,k_e)$ of $k$ for each execution. 
    Multiple executions are required to cover the entire span of interval $[0,\infty)$ until the termination condition is reached.
    
    Before describing our algorithm, we state certain properties of the FH method that lead to our interval-tracking approach and its correctness.
    Note that for a fixed input volume, all executions of the FH method go through the same list of edges and edge weights.
    Every time an edge is encountered that satisfies the condition in Equation~\ref{eqn:merging_predicate}, the two regions connected by that edge are merged. We refer to this operation as an \emph{edge collapse}.
    Felzenszwalb et al.~\cite{Felzenszwalb:2004:GBH} have proved that as long as the edges are considered in non-decreasing order of weights, 
    any order of edges will produce the same clustering for a given value of $k$.
    Nevertheless, we assume that all executions use the same fixed order of edges for the sake of simplicity.
    This condition can be enforced by pre-computing the list of edges and passing the same list to all executions. 
    \begin{property}
    \label{property:monotonic} 
    Given that the right hand side of Eq.~(\ref{eqn:merging_predicate}) is monotonic in $k$ for a fixed value of $|C_1|$ and $|C_2|$,
    during an execution of the FH method, if $k = a$ collapses an edge, all values of $k \geq a$ will also collapse that edge.
    Similarly, if $k = b$ does not collapse an edge, all values of $k \leq b$ will also not collapse that edge.
    \end{property}
    \begin{property}
    \label{property:same_regions} For any two independent executions of the FH method where $k_i \neq k_j$, 
    if both executions have made the same decisions up to an edge $e_n \in E$, then both $k_i$ and $k_j$ will encounter the same values of $|C_1|$ and $|C_2|$ for $e_{n+1}$.
    \end{property}
    \begin{property}
    \label{property:equivalence} For any two independent executions of the FH method where $k_i \neq k_j$, if all decisions to collapse edges $e \in E$ are the same,
    then the resulting clusterings of both executions are equivalent.
    \end{property}

    Our proposed exhaustive-FH algorithm repeatedly executes the FH method.
    During every execution of the FH method, we start with an interval of $k$ as $[k_s,\infty)$. Edges of the constructed adjacency graph are considered in non-decreasing order of weight. 
    Each edge is considered if it can be collapsed, i.e. if the regions connected by that edge are to be merged according to Eq.~(\ref{eqn:merging_predicate}).
    If an edge is collapsed for $k = k_s$, all values of $k > k_s$ will also collapse that edge according to Property~\ref{property:monotonic}.
    However, if an edge is not collapsed, there exists a value of $k > k_s$ that collapses the edge base on Eq.~(\ref{eqn:merging_predicate}).
    We calculate this value using the following equation:
    \begin{equation}\label{eqn:upper_limit}
        \begin{aligned}
        k_e = \text{min} \bigg( (w(C_1,C_2) - Int(C_1)) \cdot {\vert C_1 \vert}, \\ (w(C_1,C_2) - Int(C_2)) \cdot {\vert C_2 \vert} \bigg)
        \end{aligned}
    \end{equation}
    
\begin{figure}
        \centering
        \includegraphics[width=0.5\textwidth]{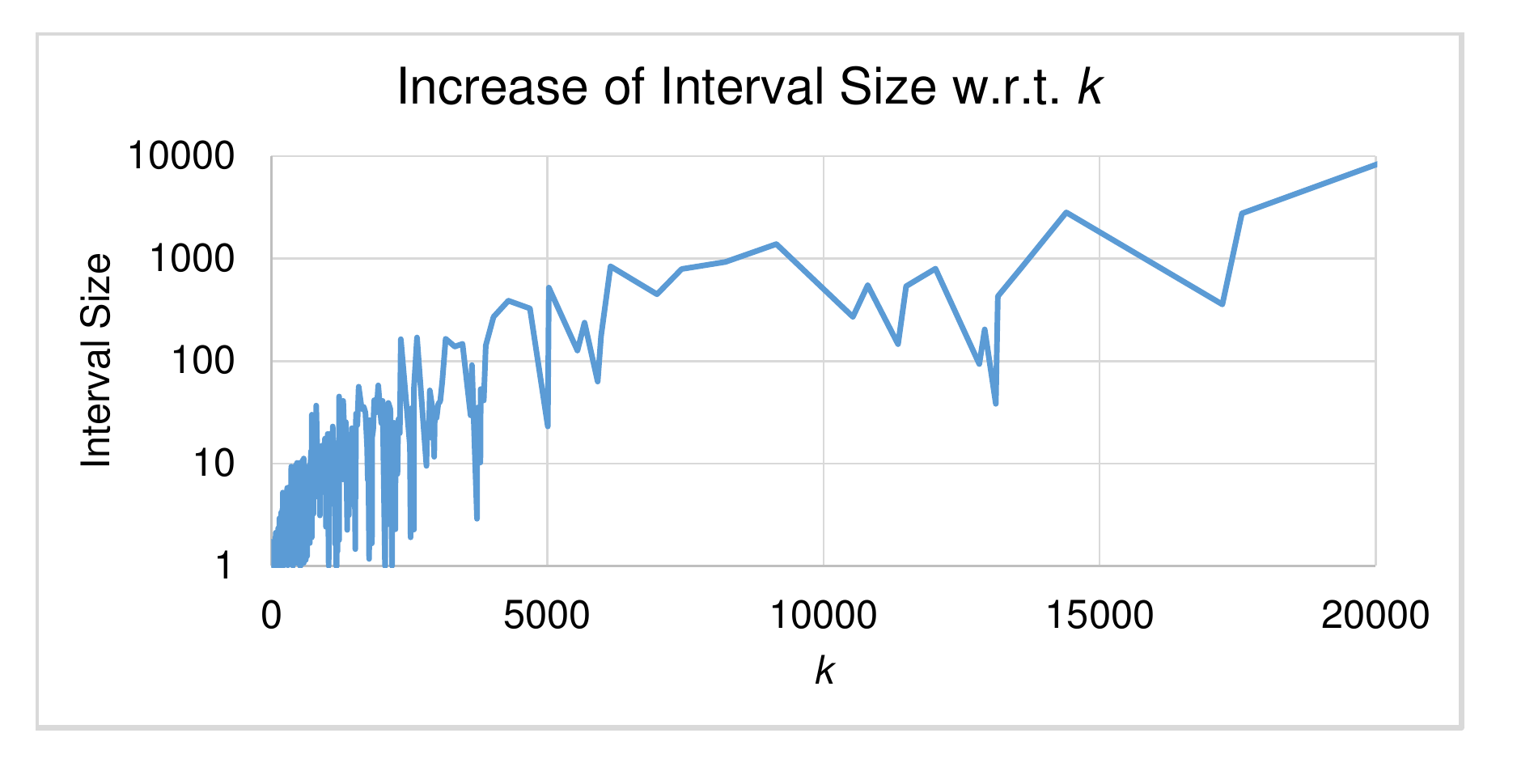}
        \vspace{-5mm}
        \caption{Graph showing the change in interval length as the value of $k$ increases for a pancreas CT dataset. 
        While the increase in interval length is not monotonic, it generally increases with higher values of $k$.}
        \label{fig:range_growth}\vspace{-3mm}
    \end{figure}

    Using the value of $k_e$, we update the tracked interval of $k$ to $[k_s, k_e)$ before the next edge in the ordered list is considered.
    Based on Property~\ref{property:monotonic} and Property~\ref{property:same_regions}, for all values of $k \in [k_s, k_e)$ at any given iteration of the FH method, we will have collapsed the same edges in $E$ and in the same order, since each edge is encountered exactly once and the order of edges is the same for all executions.
    Therefore, based on Property~\ref{property:equivalence}, we can conclude that at the end of this FH execution, all values of $k$ in the final interval $[k_s, k_e)$ will produce the same clustering.
    Every time an edge is not collapsed, this interval is updated to reflect the contiguous interval of $k$ that produces the output being generated.
    At the end of this FH execution, we have one clustering result and a contiguous interval of $k$ that produces it. 
    Any value of $k$ within this interval will produce the same clustering. 
    The next execution of the FH method uses the starting interval of $[k_e,\infty)$.
    \figurename~\ref{fig:range_growth} shows how the length of this tracked interval grows with increasing values of $k$ for a pancreas CT dataset.
    The interval size depends on the input data, but we observe in our tests that it generally increases with $k$.
    We describe our algorithm concisely in Algorithm~\ref{algo:clustering}.
    
    Edge weights in the adjacency graph represent the dissimilarity between nodes.
    Since each edge of our graph connects two super-voxels, 
    we compute the edge weight using the chi-squared distance between 1D intensity histograms of the two super-voxels.
    Each histogram uses a total of 64 bins across the entire scalar range of the input volume.
    This edge-weight policy is similar to that used by Grundmann et al.~\cite{Grundmann:2010:GBVS} for video segmentation.
    \begin{algorithm}
    \SetKwRepeat{Do}{do}{while}
        Construct compact super-voxels $B$\;
        Construct adjacency graph $G(V,E)$ for super-voxels\;
        Sort $E$ by non-decreasing order of edge-weights\;
        \textbf{Initialize} $k \leftarrow \{ 0, \infty \}$\;
        \Do{$S$.RegionCount $> 1$}{
            $S \leftarrow B$\;
            \For{\normalfont{each} $e : E$}{
                \If{$e$.weight $\leq k[0]$}{
                    Merge regions connected by $e$ in $S$\;
                }
                \Else{
                    Calculate $k_e$ using Eq.~(\ref{eqn:upper_limit})\;
                    \If{$k_e < k[1]$}{
                        $k[1] \leftarrow k_e$
                    }
                }
            }
            OutputList.insert( $S$ )\;
            $k \leftarrow \{ k[1], \infty \}$\;
        }
        return OutputList\;
        
    \caption{Exhaustive FH clustering.}
    \label{algo:clustering}
    \end{algorithm}\vspace{-3mm}

\subsection{Construction of Meta-Cluster Hierarchy}\label{sec:tree}
    Exhaustive clustering of super-voxels 
produces a large number of selectable regions.
However, proper organization of these regions is required for making them accessible to a user during the interactive exploration process.
The FeatureLego framework constructs an exploration hierarchy by grouping regions with a high degree of overlap into individual nodes of a tree.
Each node of the tree represents a group of overlapping regions that we refer to as a meta-cluster.
The meta-cluster tree reduces the time and effort to search and view similar regions. 

\subsubsection{Computing Meta-Clusters}
We compute the meta-clusters using an MST-based clustering algorithm called \emph{reverse-delete}. Given the global list of regions $R=\{r_i\}$ computed from Step 2 of our framework, a graph $G$ is constructed using regions $r_i$ as nodes and connecting each pair of regions that overlap with each other.
Edge weights are calculated using the Jaccard distance:
\[
 d_J(r_i,r_j) = 1 - \frac{|r_i \cap r_j|}{|r_i \cup r_j|}
\]
where $|r_i \cap r_j|$ is the number of voxels in the intersection of regions $r_i$ and $r_j$, and $|r_i \cup r_j|$ is the number of voxels in the union of the two regions.
The value of $d_J$ lies in the interval $[0,1]$ where $d_J = 0$ implies that $r_i$ and $r_j$ are equivalent, 
whereas $d_J = 1$ indicates that the two regions are mutually exclusive.
An MST is constructed over $G$ and edges with the highest weights are deleted repeatedly until $d_J < t$, 
where $t$ is a user-provided dissimilarity threshold.
The resulting connected components of the graph $G$ are the final meta-clusters.
In all the results, we use the dissimilarity threshold as $t = 0.3$.

\subsubsection{Tree Construction}
We construct the meta-cluster tree $T(V,E)$, where each node $v \in V$ is a meta-cluster as computed in Step 3 of the FeatureLego pipeline.
For the convenience of notation in this text, we refer to the tree nodes and meta-clusters interchangeably.
The edges $E$ between nodes $V$ are constructed based on the containment relationship among the meta-clusters.
For each node $v_c$, a parent node $v_p$ is found as the smallest meta-cluster that is a superset of $v_c$.
Any node for which a parent cannot be found is directly connected to the root node, since there is no other node that contains it. 
The containment constraint for the parent-child relationship in $T$ is still respected as the root node represents the entire input volume.
\begin{figure}[!t]
    \centering
    \includegraphics[width=0.47\textwidth]{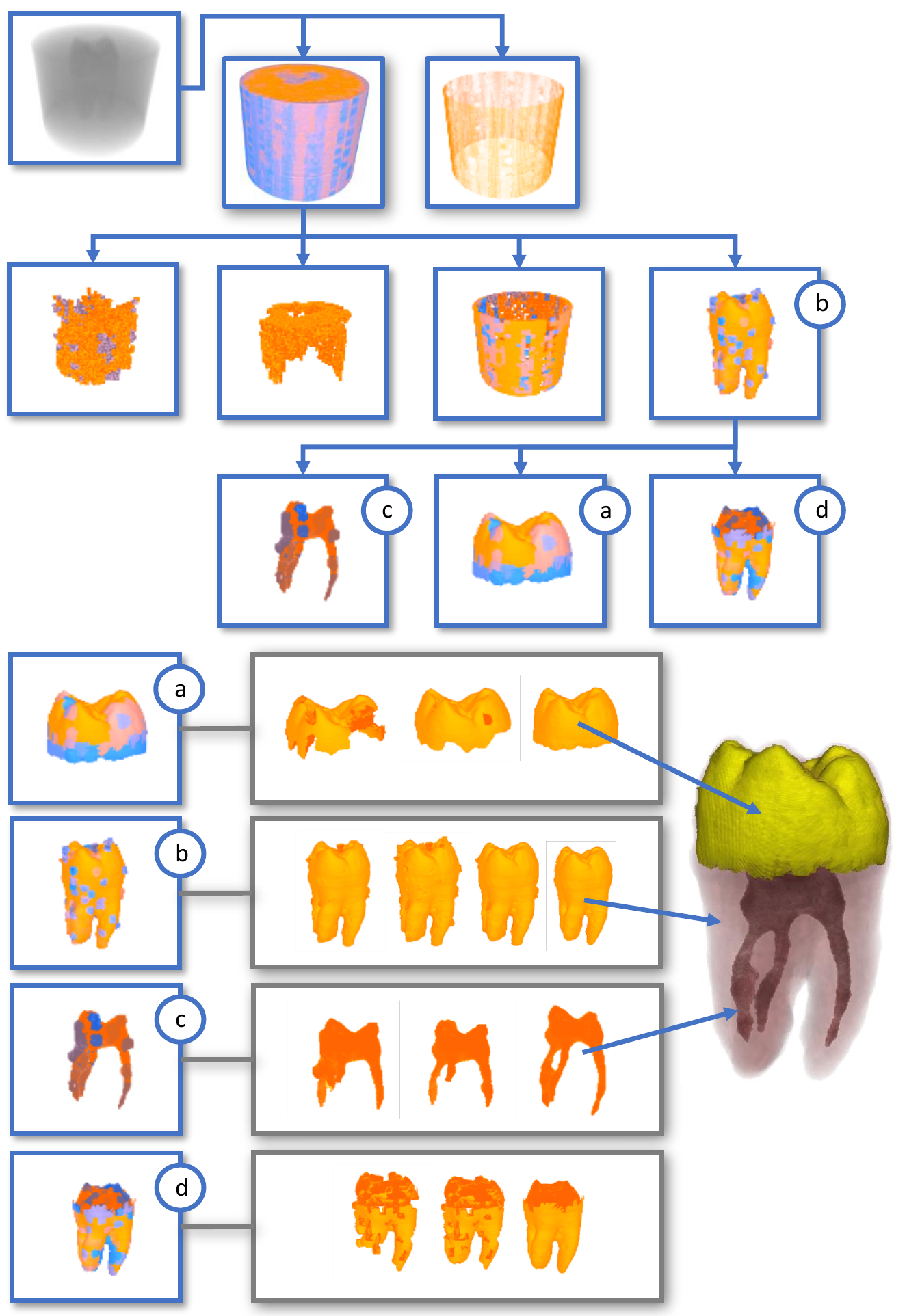}
    \caption{Partial meta-cluster tree for the Tooth dataset. The root node represents the entire dataset.
    Each non-root node of the tree represents a cluster of regions (meta-cluster). 
    Nodes (meta-clusters) with expanded members are shown at the bottom. 
    A user selects individual regions from the expanded meta-clusters to construct a visualization.
    }\vspace{-5mm}
    \label{fig:tree}
\end{figure}
Additionally, to ensure that all nodes $v_s$ that contain a node $v_c$ are ancestors of $v_c$ in $T$, 
we insert duplicate copies of $v_c$ under every node $v_s$, if $v_s$ is not already an ancestor to $v_c$.
Some supersets may already be ancestors of other supersets.
In such a case, adding $v_c$ to the smaller superset is sufficient,
since the larger superset will share the same copy of $v_c$ under their common branch.
Hence, the insertions are made in the order of increasing sizes of the superset nodes to ensure that
duplicate instances of $v_c$ in the tree are minimized.
Additionally, each set of sibling nodes is sorted by decreasing size in voxels. 
Consequently, the user is intuitively guided regarding when to stop or change direction in a linear search through a set of sibling nodes,
based on the expected size of a semantic feature.
This also helps in dynamic filtering of the tree using user-specified minimum meta-cluster size and maximum branching factor.
We discuss the details of the dynamic filtering ability in section~\ref{sec:ui}.
\begin{figure}
    \centering
    \includegraphics[width=0.47\textwidth]{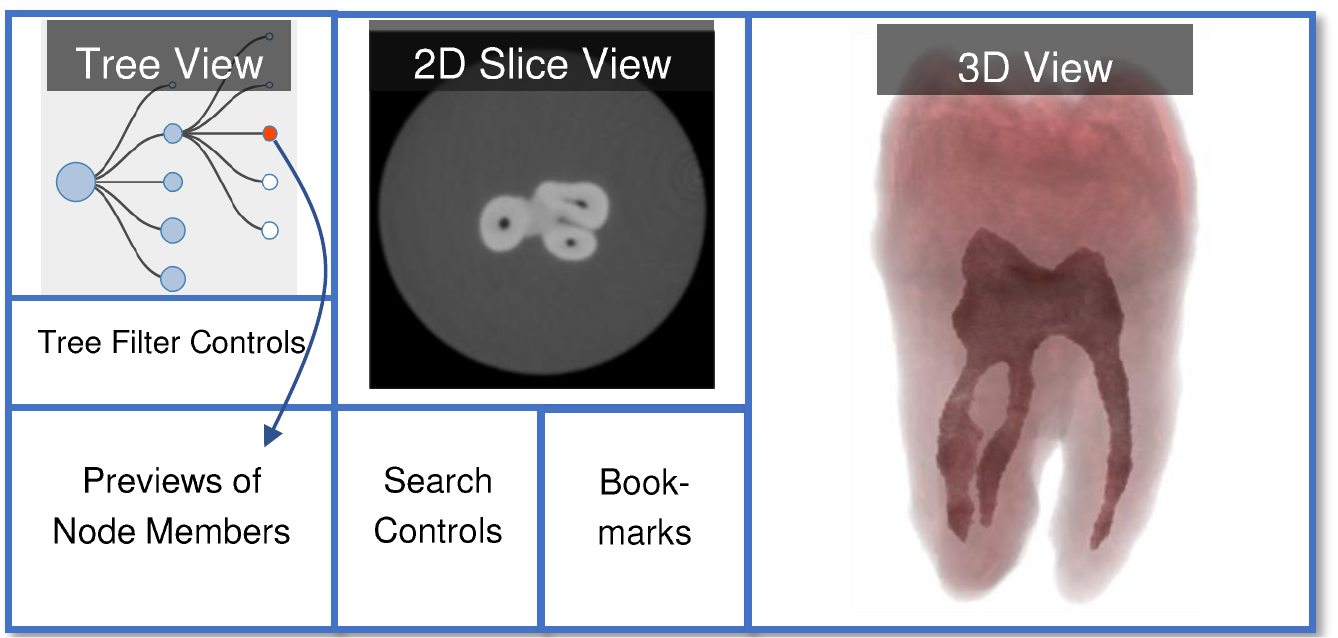}
    \caption{Outline of the exploration interface.
            \emph{Tree View} renders the meta-cluster tree and previews of regions of a selected node are shown in the \emph{Node Members} view.
             \emph{Tree Filter Controls} are used to dynamically prune the tree based on minimum meta-cluster size and maximum branching factor. 
             Selected regions are saved to \emph{Bookmarks} list view where optical properties can also be modified for each bookmark.
             \emph{Search Controls} are used for searching features through brushing and size constraints.
             }\label{fig:ui_outline}\vspace{-4mm}
\end{figure}
It is important to note that the region sizes used for finding parent nodes and in calculating $d_J$
are measured in terms of voxels and not in terms of super-voxels. This is important to quantify the
actual sizes of regions in the volume.
A detailed description of the tree construction algorithm is provided in Algorithm~\ref{algo:tree}.
\begin{algorithm}
\SetKwRepeat{Do}{do}{while}
    Compute meta-clusters $\{v_i\}$\\
    \For{\normalfont{each} $v_i$}{
        Construct superset list $\{s_j\}$ such that $v_i \subseteq s_j$\\
        Sort superset list by increasing sizes\\
        \If{$\{s_j\}$.NotEmpty}{
            Find smallest superset $s_f$\\
        }
        \Else{
            $s_f = root$\\
        }
        Construct edge $(v_i,s_f)$\\
    }
    \For{\normalfont{each} $v_i$}{
        \For{\normalfont{each superset} $s_j$}{
            \If{$s_j$ \normalfont{is not an ancestor of} $v_i$ }{
                Create node $v'_i$ as duplicate of $v_i$\\
                Construct edge $(v'_i,s_j)$\\
            }
        }
    }
    \For{\normalfont{each} $v_i$}{
        Sort child nodes by size in descending order.
    }
\caption{Meta-cluster tree construction.}
\label{algo:tree}
\end{algorithm}

\begin{figure*}[ht]
    \centering
    \includegraphics[width=0.95\textwidth]{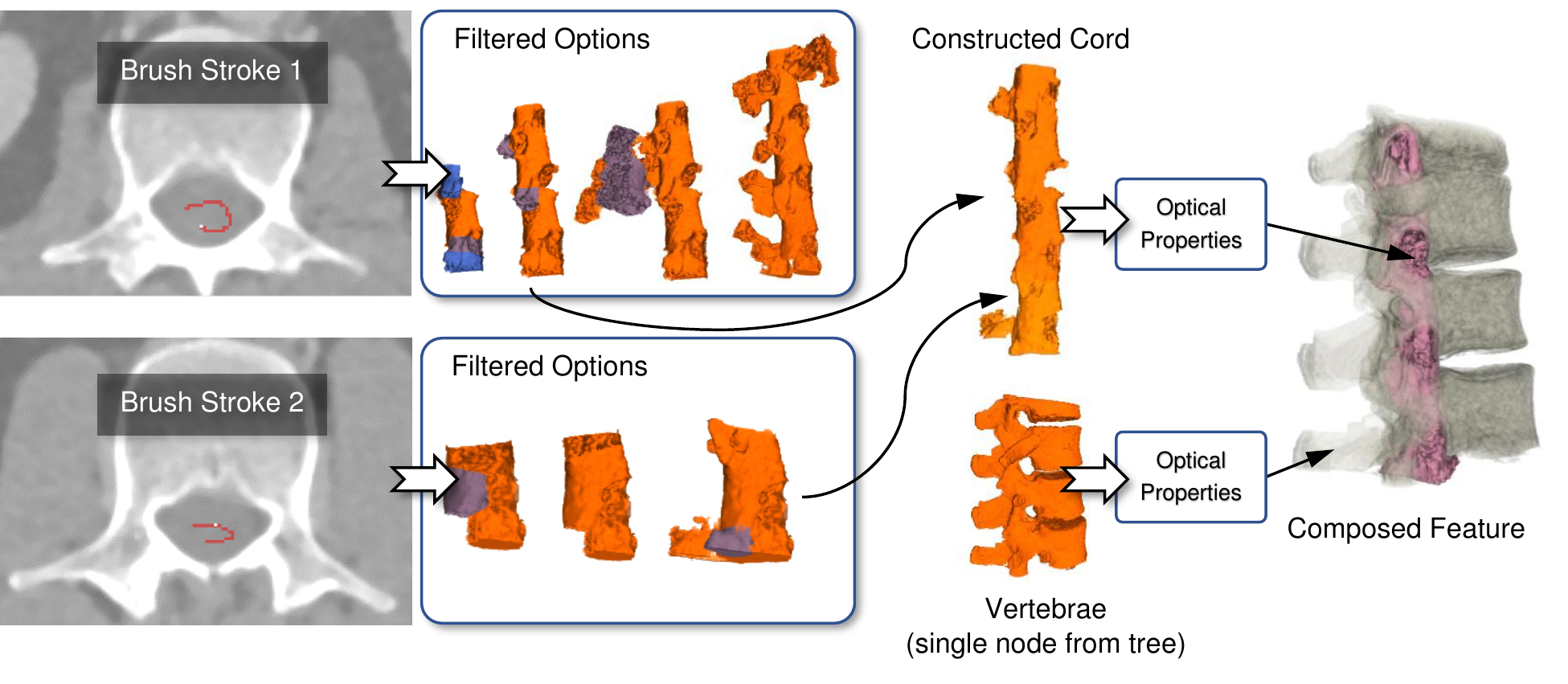}
    \caption{Illustration of a workflow for selecting complex features in the user interface. 
    Two brush strokes along with the search operation are used to find the two pieces of the spinal cord. 
    A minimum size of 1000 and maximum size of 100,000 were used in the search queries.
    Vertebrae are selected directly as a single region from the tree.}
    \label{fig:brushing_illustration}\vspace{-3mm}
\end{figure*}

An example of the meta-cluster tree for the Tooth dataset is shown in \figurename~\ref{fig:tree}.
The figure shows a partial exploration tree where a blue-orange transfer function is used 
to depict the number of overlapping regions of the meta-cluster in each voxel.
A blue voxel indicates fewer overlapping regions, whereas an orange voxel represents a large number of overlapping regions.
High opacity is used so that a user can correctly judge the differences between boundaries of overlapping regions. Original scalar intensities from input volume are not used in this type of rendering.
As shown in the figure, the root node contains the entire input volume and the first level nodes provide 
a broad separation between the region-of-interest (the tooth inside the cylinder) and the surrounding noise.
Since we construct the exploration tree such that smaller regions that are contained within larger regions 
are guaranteed to be found as descendant nodes, the user can simply continue down the tree by selecting
the cylinder node and ignore all other nodes.
Expanding the cylinder node reveals node (b) that contains the entire tooth.
Further expansion of node (b) separates the dentine, root canals, and the crown.
An expanded list of regions belonging to each of nodes (a), (b), (c), and (d) are shown
below the tree.
The user selects appropriate regions from the tree-nodes to construct meaningful visualization of the tooth.

We also show the exploration of the Tooth dataset using the meta-cluster tree in the supplementary video.
The previews shown for each node are not actually embedded in the tree-view of our user-interface.
They are shown in the figure for illustrative purposes. Details of the user interface are discussed in Section~\ref{sec:ui}.


\subsection{User Interface and Implementation}
    \label{sec:ui}
We implement an efficient user interface to filter and explore the meta-cluster tree, to select and bookmark semantic features,
and to edit optical properties of features to generate meaningful visualization of volumes.
An outline of our interface is given in \figurename~\ref{fig:ui_outline}.
An interactive collapsible graph rendered in D3.js is used for navigating the meta-cluster tree. We achieve this by embedding Qt's web engine in our C++ application.
The nodes are rendered as solid circles of variable size based on the size of the meta-clusters that they represent. This provides the user with size hints during exploration.
A node can be selected using a single-click and it can be expanded using a double-click.
If a node is selected, its individual member regions (regions clustered together as a single node of the tree) are shown in a list view below the tree.
These are shown as pre-rendered snapshots of each region, so that the user can quickly find interesting features.
Nodes containing noisy regions with high-intensity voxels refuse to merge with neighboring regions until very high values of $k$. Most noisy regions can be removed through initial smoothing of the input volume, however,
many of these noisy regions create a large fan-out closer to the root node.
We provide the user the ability to dynamically filter the tree based on specified minimum meta-cluster size (in voxels) and maximum branching factor to overcome high branching during exploration.
This capability can be used to simplify the exploration process if the user knows the size of features they want to search.
Pruning the tree by maximum branching factor has the effect of eliminating the smallest siblings in each set of child nodes, since all siblings are sorted by decreasing order of size.

Filter controls are used to execute search queries over the meta-cluster tree.
First, a brushing tool is used on the 2D slice view to select voxels of the input volume.
Then, the user specifies the minimum and maximum meta-cluster sizes 
in provided text boxes.
Using the selected voxels and the size constraints, the filtered nodes (meta-clusters) are shown in a list-view.
During the execution of the search query, first the super-voxels that contain all the selected voxels are identified. The leaf nodes of the meta-cluster tree are super-voxels computed in Step 1 of our pipeline.
A bottom-up search for nodes is performed on the tree based on the identified leaf nodes.
All nodes (meta-clusters) that match the minimum and maximum size constraint are identified. This query can terminate early based on the maximum size constraint because
the parent nodes are guaranteed to be equal or larger than the child nodes.
The range of minimum and maximum sizes for search queries is guessed intuitively by the user, and does not need to be precise.
In our experiments, we found that entering values based on orders of magnitude (powers of ten) is sufficient to find the desired features
and helps in terminating the search query quickly.
We implement linking of views so that the user can jump to a corresponding tree node by double-clicking an item in the filtered results,
or by brushing and double-clicking the 2D slice view. Double-clicking the 2D view will jump the tree to the smallest meta-cluster node that completely contains the brushed voxels.

An illustration of the workflow is shown in \figurename~\ref{fig:brushing_illustration} 
for selection of vertebrae and spinal cord in an Abdominal CT dataset. This same example
is also demonstrated in the supplementary video.
The brushing and search tool can be used specifically for small features that may be harder to find in the tree view.
It is also helpful when adding missing pieces in a bigger semantic feature.

Selected regions can be saved to bookmarks for later reference.
A user constructs meaningful visualizations by separating out features of interest as bookmarks.
Each bookmark item can be visualized independently or as a combination with other bookmarks.
An initial 1D transfer function is automatically generated for each bookmarked feature, which provides the user with a good starting point to further modify the optical properties.
We generate the 1D transfer function automatically by computing a one-dimensional intensity histogram as a curve (polyline) and simplify it using persistence-based topological simplification~\cite{Gyulassy:2005:3DMS} to identify its most prominent ``bumps" or features.
Colors are then applied to the points of this simplified polyline using diverging palettes from \emph{ColorBrewer}~\cite{harrower2003colorbrewer}.
The interface provides three rendering modes for bookmarked features: surface rendering, flat rendering,
and one-dimensional transfer function.
These rendering modes can be used in different combinations for reducing
occlusions and improving the overall visualization when visualizing multiple features.

We implement the super-voxel computation in Step 1 of our pipeline, by extending the GPU implementation of SLIC provided by Ren et al.~\cite{gSLICr_2015} to 3D volumes.
We implement the exhaustive-FH clustering algorithm using multi-threading.
Each thread is passed a range of $k$ for processing which we refer to as processing-range.
Each thread runs the FH clustering algorithm multiple times to identify maximal contiguous intervals of $k$ and the corresponding clusterings within the passed processing-range.
Based on \figurename~\ref{fig:range_growth}, we expect smaller contiguous intervals when values of k are closer to zero. This means that a thread will require more iterations of FH method to process a range closer to zero than higher values.
Therefore, for balancing the workload, we pass these processing-ranges to threads in increasing order and with increasing size.
For example, we use a range growth factor of 1.5 so that each consecutive thread gets a range that is larger by a factor of 1.5, i.e. thread1 gets a processing-range of [0, 50), thread2 gets [50, 125), and so on.
Clustering results are written to separate output files as they are generated in each iteration.
A thread terminates after its processing-range of $k$ has been exhausted.
We maintain a fixed number of threads at any time. 
If a thread terminates, a new thread is started with a processing-range that has not been processed before.
If any thread terminates with the number of regions equal to 2 or less, further instantiation of threads is stopped,
since higher values of $k$ are not going to produce any new clusterings.
We use 12 simultaneous threads in all our experiments.
Once all the clustering threads are finished, the main process starts the meta-cluster tree construction.
The generated output files are then processed to identify unique regions and grouped into meta-clusters.
Meta-clusters are then constructed into a tree using Algorithm~\ref{algo:tree} as described in Section~\ref{sec:tree}.

\vspace{-1mm}
\section{Results and Evaluation}
    We evaluate the capabilities of the FeatureLego framework in two parts: qualitative and performance evaluation.
The qualitative evaluation is performed by comparing (a) previous voxel clustering approach by Ip et al.~\cite{Ip:2012:hierarchical},
(b) FeatureLego with parameter sampling, and (c) FeatureLego with exhaustive clustering.
We also demonstrate our framework's superior capabilities to separate features in complex real-world datasets.

\begin{figure*}[!ht]
\centering
\includegraphics[width=0.95\textwidth]{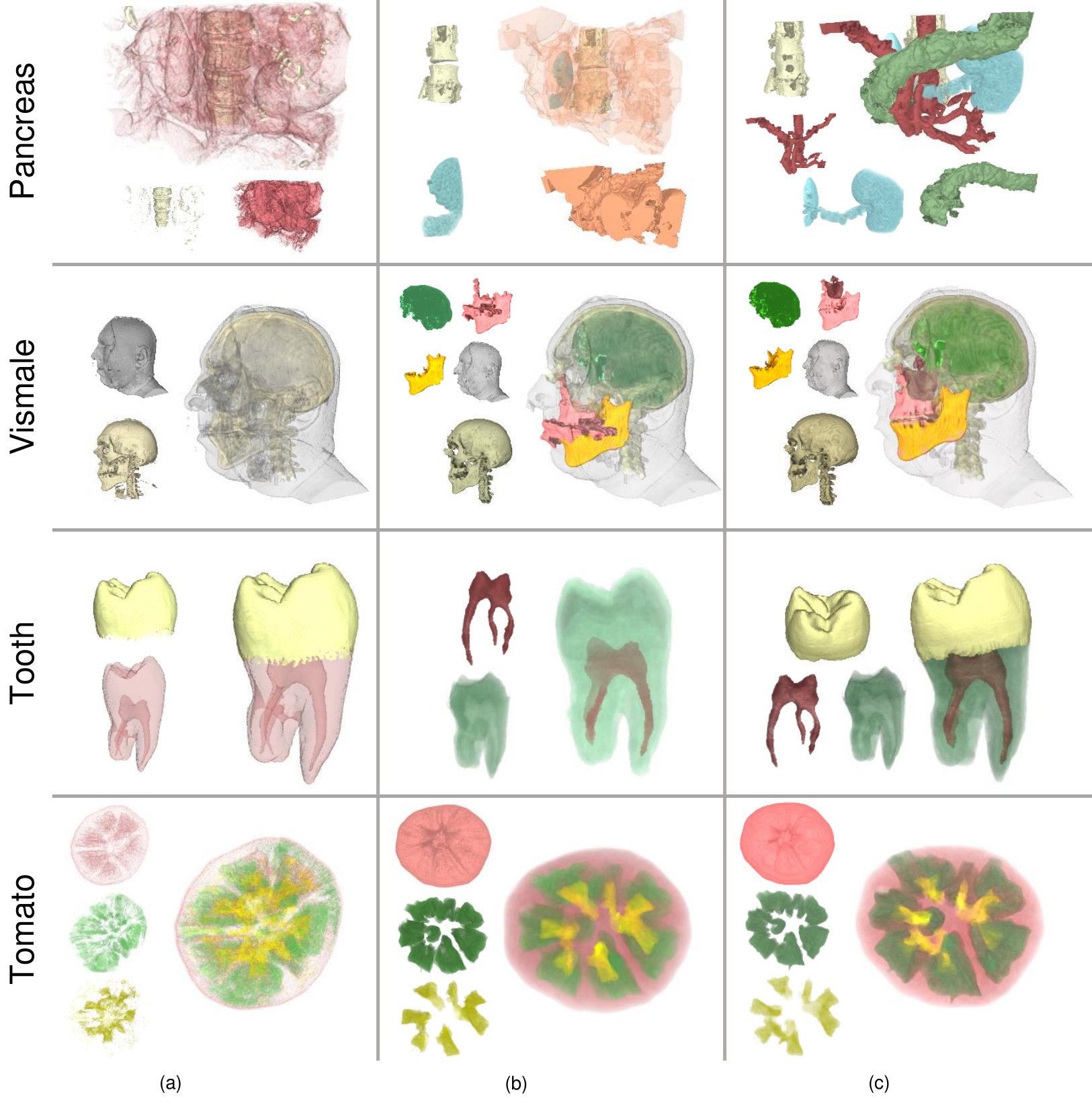}
\caption{Comparison between (a) voxel-clustering through normalized-cut of 
intensity-gradient histograms, (b) FeatureLego with parameter sampling, and (c) FeatureLego with exhaustive clustering.
Datasets shown in each row from top-to-bottom are Pancreas, Visible Male Head, Tooth, and Tomato.
The histogram normalized-cut method is limited in separating features and features can often be fragmented.
Parameter sampling misses out on some features, while exhaustive clustering is able to separate more semantic features than the other two methods.
}\label{fig:compare}\vspace{-3mm}
\end{figure*}

\subsection{Qualitative Evaluation}
\figurename~\ref{fig:compare} shows a comparison between the aforementioned techniques in three columns. 
The first column displays results from the normalized-cut method on intensity-gradient histograms (ncut); 
the second column shows results from FeatureLego with parameter sampling; and, the third column
shows results from FeatureLego with exhaustive clustering.
Histogram segmentation was performed as a binary tree up to six levels, 
and parameter sampling results were generated using uniform sampling of $k$ with interval size equal to 5000.
This interval size is chosen arbitrarily and it represents a guess that the user has to make to find features.
The user may have to recompute results with different values until the desired result is achieved.
A smaller interval size may capture more features, but it is impossible to know whether a desired feature can be separated at all.
On the other hand, our exhaustive clustering algorithm does not require this guessing and computes all possible clusters in a single run.

Each separated feature is rendered using a single opacity and color value for easy comparison.
Rows show Pancreas, Vismale, Tooth, and Tomato datasets from top to bottom. 
Details of these datasets are shown in Table~\ref{tab:data_size}.
\begin{table}[!t]
    \centering
    \caption{Dimensions and scalar range of datasets used in our tests, sorted by voxel count.}
    \label{tab:data_size}
    \begin{tabular}{|c|c|c|c|}
        \hline
        \textbf{Dataset} & \textbf{Dimensions} & \textbf{Scalar Range} & \textbf{Modality} \\
        \hline 
         Tomato   & $256 \times 256 \times 64$  &    0 to 255        & MRI \\
         Pancreas & $235 \times 153 \times 210$ & -854 to 1087       & CT \\
         Vismale  & $128 \times 256 \times 256$ &    0 to 255        & CT \\
         Tooth    & $256 \times 256 \times 161$ &    0 to 1300       & CT \\
         Knee MRI & $512 \times 512 \times 120$ &   15 to 3668       & MRI \\
         Chest CT & $384 \times 384 \times 240 $ & 0 to 255          & CT \\
         Abdominal CT & $504 \times 416 \times 243 $ & -1024 to 1311 & CT \\
        \hline
    \end{tabular}
\end{table}

\textbf{Pancreas.} Segmentation of the 2D intensity-gradient histogram is unable to separate any of the abdominal organs including the pancreas.
It is effective in separating boundaries but not the individual semantic features due to overlapping intensities
and lack of spatial information.
Further subdivision results in fragmented regions without clear separation of features.
The FeatureLego pipeline with parameter sampling separates the vertebrae and one kidney but is unable to separate other organs.
It groups all the remaining features into a single node. Subsequent child nodes are the smallest granularity super-voxels.
FeatureLego with exhaustive clustering is able to separate many important features such as the vertebrae, kidneys, pancreas, and
partial vascular structure. 

\textbf{Vismale Head.} This is a CT scan of the Visible Male Head. The histogram segmentation approach is able to cleanly separate the tissues and
bones. However, it fails to provide clean separation of features within the tissues. 
Further segmentation down the tree only leads to fragmented nodes without clear separation of compact regions.
FeatureLego with parameter sampling is able to separate some features but does not separate the jaw correctly.
Denser sampling of $k$ may provide correct separation.
FeatureLego with exhaustive clustering is able to separate the upper and lower jaw, skull, skin, and brain.

\textbf{Tooth.} The histogram segmentation approach works quite well. It is able to separate important boundaries of the root
canals and the crown. The parameter sampling approach is unable to separate the crown, whereas exhaustive clustering
separates the crown, the root canal, and the dentine. It separates them as regions and not just the boundaries.

\textbf{Tomato.} FeatureLego with sampling as well as with exhaustive clustering performs comparably on this dataset.
Histogram segmentation is also able to highlight important features, but the regions are more fragmented.

\begin{figure}[t]
    \centering
    \includegraphics[width=0.8\linewidth]{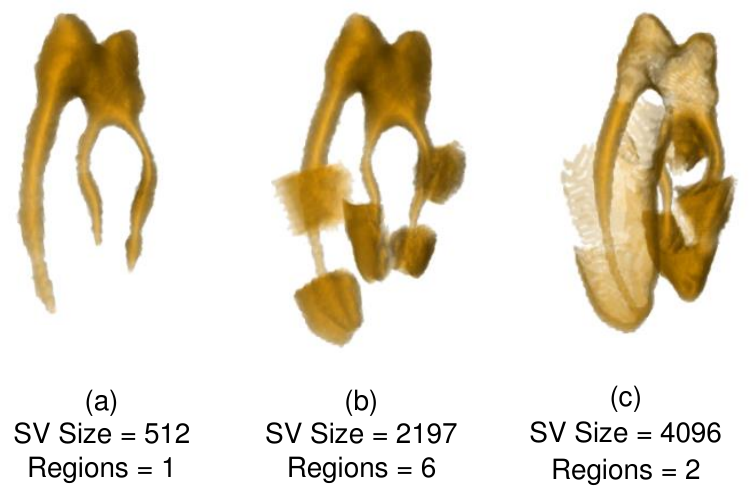}
    \caption{Effect of changing super-voxel size (base granularity) on Tooth dataset.
    Each sub-figure shows super-voxel size used and number of regions combined to select the root canals.
    The root canals are impossible to separate cleanly (b and c) until super-voxel size is lowered to 512 (a).}
    \vspace{-2mm}
    \label{fig:granularity_effect}
\end{figure}

We tested the effect of base granularity (size of SLIC super-voxels) on the selection of semantic features. 
The result for selecting the root canals in the Tooth dataset is shown in \figurename~\ref{fig:granularity_effect}.
Super-voxel sizes of 4096 and 2197 do not produce exact separation of the root canals. 
Additionally, the root canals had to be assembled through multiple regions.
On the other hand, for a smaller super-voxel size (512), the root canals are separated cleanly and can be selected
as a single region from the meta-cluster tree.
Due to the fine geometry of the canals, a smaller super-voxel size had to be chosen to separate them correctly.
This is a limitation of our framework as the user has to select an appropriate super-voxel size based on the size of semantic features.
Nevertheless, we found that a super-voxel size of 2197 and 4096 worked for most input datasets that we tested.
Specifically, we use a size of 512 for Tooth and Vismale dataset; 2197 for Tomato and Pancreas; and, 4096 for Knee MRI, Chest CT and Abdominal CT data.

\begin{figure*}
\centering
\includegraphics[width=0.95\textwidth]{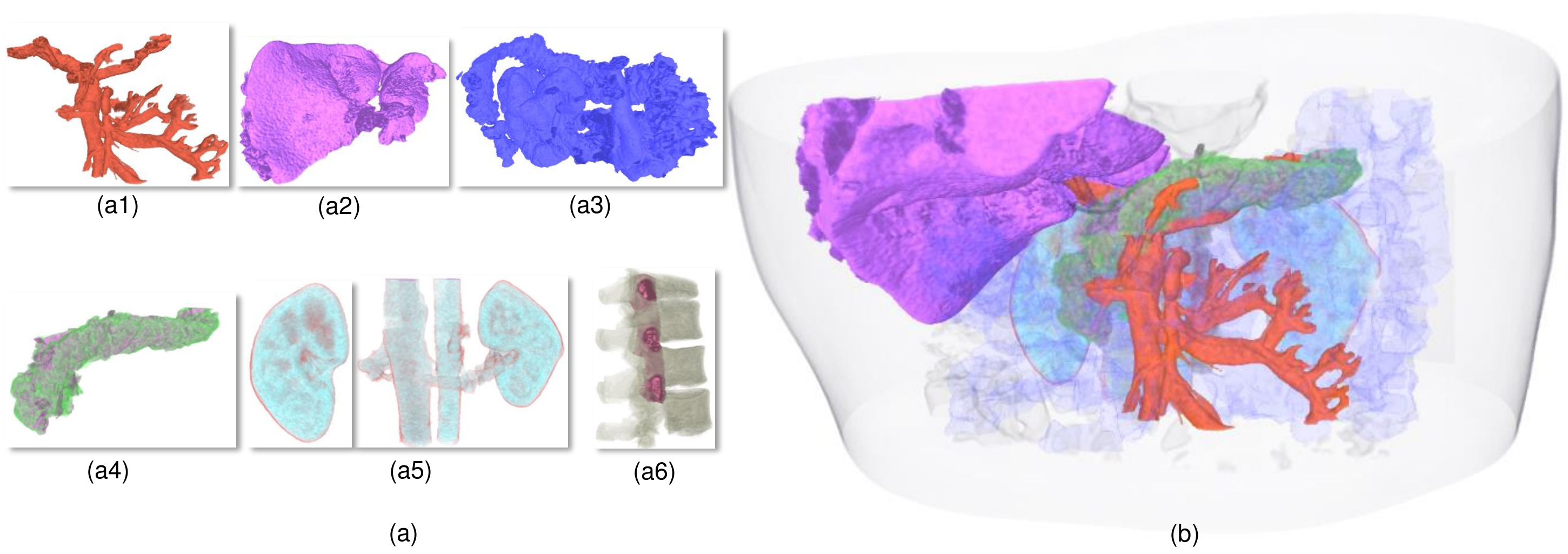}
\caption{ Abdominal CT scan. (a) Individually separated features, and (b) combined visualization.
Separated semantic features include (a1) partial vascular structure, (a2) liver, (a3) colon, (a4) pancreas, (a5) kidneys, (a6) vertebrae and spinal cord.}
         \label{fig:vp11_roi}
\end{figure*}

\begin{figure*}
\centering
\includegraphics[width=0.95\textwidth]{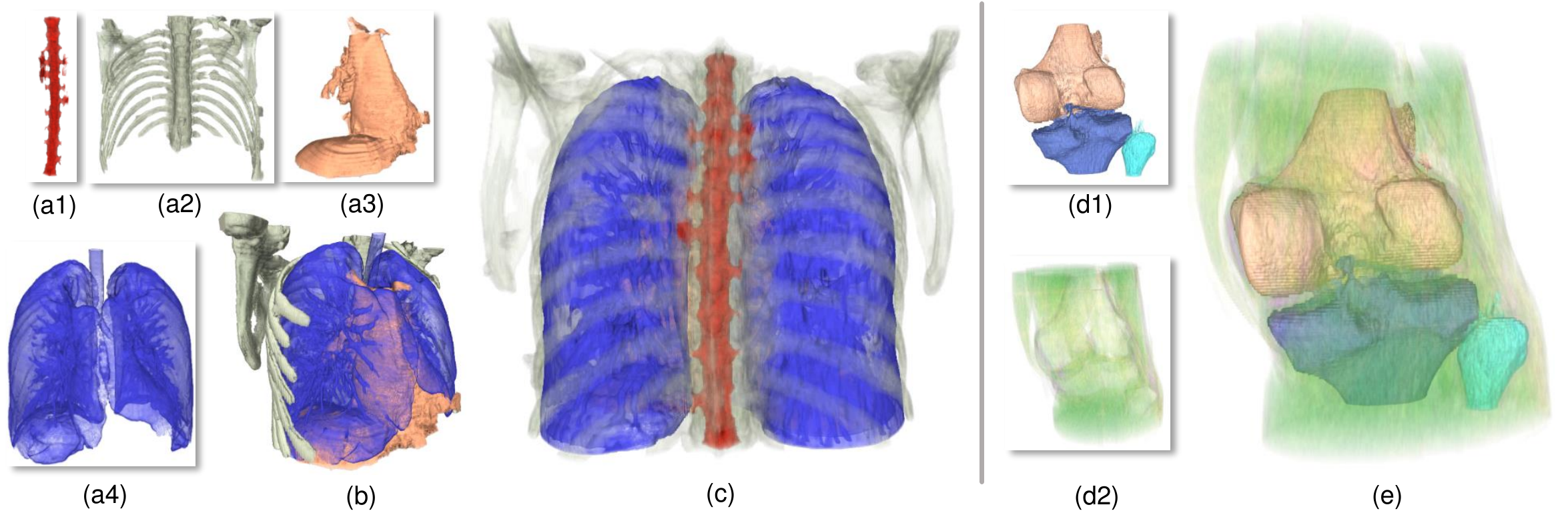}
\caption{Chest CT and Knee MRI scans. Separated features for the Chest CT (a) along with two views (b-c) of the composed visualization.
Semantic features include (a1) spinal cord, (a2) rib cage, (a3) heart and other partial organs, and (a4) lungs.
Composed visualization is shown at two different viewing angles: (b) side view, and (c) rear view of the volume.
Semantic features for the Knee MRI include (d1) bone structure - femur (orange), tibia (blue), and fibula (cyan), (d2) tissue region.
Composed visualization of the Knee MRI dataset is shown in (e). } 
\vspace{-5mm}
\label{fig:chest_ct}
\end{figure*}

Results on additional real-world datasets are shown in \figurename~\ref{fig:vp11_roi} and \ref{fig:chest_ct}.
The dataset used in \figurename~\ref{fig:vp11_roi} is an abdominal CT scan from which the Pancreas dataset was extracted using a bounding box around the pancreas.
In the Abdominal CT, we are able to separate the colon, both kidneys, pancreas, liver, vertebrae, spinal cord, and partial vascular structure
within the abdomen. 
The dataset used in \figurename~\ref{fig:chest_ct} is a chest CT scan.
In this data, our framework is able to separate important features such as the spinal cord, the rib cage, heart and other organs, and lungs.
These features are then composed into a single visualization with two different view points.
Table~\ref{tab:selection} shows how each semantic feature was selected for the two real-world datasets. It can be observed that some features require only
a single selection of a node within the exploration tree, whereas some have missing fragments that need to be selected through brushing-based filtering.
The workflow for selecting semantic features by combining regions from the exploration tree and brushing-based 
filtering is illustrated in \figurename~\ref{fig:brushing_illustration}.

In qualitative evaluation, we showed results from a total of 7 datasets.
Following are the approximate time durations we took for constructing each of these visualizations including separating features and editing optical properties.
Tooth, Pancreas, and Tomato datasets took 2, 6, and 12 minutes respectively. 
Vismale, Knee, and Chest CT each took around 15 minutes.
Whereas, Abdominal CT took around 18 minutes for constructing the visualization.
\begin{table}[!t]
    \centering
    \caption{Number of regions and selection method for semantic features in real-world datasets (Figs.~\ref{fig:vp11_roi}~and~\ref{fig:chest_ct}).}
    \label{tab:selection}
    \begin{tabular}{|c|c|c|c|c|}
        \hline
        \textbf{Dataset} & \textbf{\makecell{Semantic\\Feature}} & \textbf{\makecell{Selected\\Through\\Tree}} & \textbf{\makecell{Selected\\Through\\Brushing}} & \textbf{\makecell{Total\\Regions}} \\
        \hline 
                  & Spinal cord (a1) & 1 & 1 & 2 \\ 
         Chest CT & Rib cage (a2) & 1 & 9 & 10 \\ 
                  & Heart region (a3) & 1 & 11 & 12 \\ 
                  & Lungs (a4) & 1 & 0 & 1 \\ 
         \hline 
                  & \makecell{Vessels (a1)} & 2 & 0 & 2 \\ 
                  & Liver (a2) & 1 & 0 & 1 \\ 
        Abdominal & Colon (a3) & 1 & 0 & 1 \\ 
           CT     & Pancreas (a4) & 1 & 4 & 5 \\ 
                  & Kidneys (a5) & 2 & 0 & 2 \\ 
                  & \makecell{Vertebrae (a6)} & 1 & 2 & 3 \\ 
        \hline
                  & Femur (d1) & 1 & 6 & 7 \\ 
         Knee MRI & Tibia (d1) & 1 & 8 & 9 \\ 
                  & Fibula (d1) & 0 & 1 & 1 \\ 
                  & Tissue (d2) & 1 & 0 & 1 \\ 
         \hline 
    \end{tabular}
\end{table}
\begin{table}[!t]
    \centering
    \caption{Pre-processing times required for each dataset at different stages of the pipeline.}
    \label{tab:time}
    \begin{tabular}{|c|c|c|c|c|}
        \hline
        \textbf{Dataset} & \textbf{SLIC} & \textbf{\makecell{Exhaustive\\Clustering}} & \textbf{\makecell{Tree\\Construction}} & \textbf{\makecell{Total\\Time}} \\
        \hline 
         Tomato   & 17 sec  & 17 sec & 3 sec & 37 sec \\
         Pancreas & 30 sec  & 38 sec & 22 sec & 1.5 min \\
         Vismale  & 33 sec  & 2.1 min & 4.4 min & 7 min \\
         Tooth    & 42 sec  & 74 sec & 86 sec & 3.4 min \\
         Knee MRI & 2  min  & 7.3 min & 7.9 min & 17.2 min \\
         Chest CT & 2.4 min & 4.1 min & 4.8 min & 11.3 min \\
         Abdominal CT & 3.4 min & 6.7 min & 10.2 min & 20.3 min \\
        \hline
    \end{tabular}
\end{table}
\begin{table}[ht]
    \centering
    \caption{Meta-cluster tree reduces the number of elements to be explored by a user, 
    by removing duplicates and grouping overlapping regions.}
    \label{tab:tree}
    \begin{tabular}{|c|c|c|c|c|}
        \hline
        \textbf{Dataset} & \textbf{\makecell{Regions\\(Million)}} & \textbf{\makecell{Meta-\\Clusters}} & \textbf{\makecell{Tree\\Nodes}} & \textbf{\makecell{Memory\\(GB)}} \\
        \hline 
         Tomato   & 1.1 & 3,271 & 5,496 & 0.3\\
         Pancreas & 5.2 & 6,380 & 11,792 & 1\\
         Vismale  & 39 & 28,786 & 121,868 & 1.3\\
         Tooth    & 10.4 & 18,978 & 60,834 & 4.7\\
         Knee MRI & 86 & 34,194 & 62,329 & 13.7\\
         Chest CT & 40.3 & 17,158 & 37,899 & 6.8\\
         Abdominal CT & 117.7 & 39,644 & 127,332 & 23\\
        \hline
    \end{tabular}
\end{table}

\vspace{-4mm}
\subsection{Performance Evaluation}
All results were generated on a desktop workstation with an Intel Xeon E5-2620 processor (6 cores, 12 threads), 32 GB main memory,
and Nvidia GTX 1080 Ti graphics card.
Table~\ref{tab:data_size} shows all the test datasets sorted by size (voxel count) and
Table~\ref{tab:time} shows the pre-processing time required at each stage. 
These timings depend upon the size of the volume as well as the number of super-voxels computed by SLIC.
The SLIC process grows slower with larger input volumes as observed in Table~\ref{tab:time}, but can be faster if using a larger super-voxel size.
The exhaustive clustering and meta-cluster tree construction times mostly depend upon the number
of super-voxels computed by the previous step, since all further computations happen over
the level of super-voxels.


Exhaustive clustering creates regions as different combinations of initially computed super-voxels from Step 1 of our framework. This creates a set of regions that overlap with each other including duplicate regions at different intervals of $k$.
Table~\ref{tab:tree} shows the reduction of the total number of initially computed regions into nodes of the meta-cluster tree based on the removal
of duplicates and clustering of overlapping regions. All trees were constructed using a dissimilarity threshold of $t = 0.3$.
Column 2 shows initial regions computed during exhaustive clustering. Column 3 shows the total meta-clusters computed after removing duplicate regions and clustering overlapping regions into meta-clusters.
After construction of the meta-cluster tree and repeating nodes
to ensure containment, the node count increases slightly (Column 4).
A tree with thousands of nodes may seem large. However, the user focuses on specific regions of interest that narrow the search to certain branches of the tree.
Memory required for processing and exploring each dataset is shown in Column 5.
Memory consumption increases due to required data-structures for supporting search queries and caching during exploration.
\vspace{-2mm}

\section{Conclusion and Future Work}
    In this paper, we have presented an exhaustive super-voxel clustering pipeline 
to efficiently and intuitively select semantic features for visualization of volumetric data.
The process of computing selectable regions and organizing them for exploration is performed as an offline pre-processing step.
A user can then browse through our meta-cluster tree to select regions of interest in order to construct meaningful visualizations. 
An initial base granularity is used for enforcing local compactness and efficiency during the clustering process.
By performing exhaustive clustering, we provide the user with more choices when separating semantic features without the need
to re-execute the clustering algorithm with different input parameters.
We presented an efficient algorithm to perform exhaustive clustering using the FH clustering method~\cite{Felzenszwalb:2004:GBH}; and, described a technique to construct an exploration hierarchy
using overlap and containment between segmented regions.
We implemented an intuitive exploration interface for searching, bookmarking, and visualizing semantic features.
Finally, we demonstrated the capabilities of our FeatureLego framework on complex real-world datasets.

Based on voxel count, we have tested the performance of our framework for volumes of size up to $370^3$.
For future work, we aim to scale our framework to significantly larger volumes such as microscopy images.
The tree construction and exploration steps of our pipeline use a significant amount of memory, 
which may be alleviated by developing more efficient data-structures and utilizing compression to represent voxel clusters.
Such a representation should allow for efficient search queries and set operations while
maintaining a low memory foot-print.
Moreover, we chose the FH method as it has properties that can be used for interval-tracking of its input parameter.
A similar approach can be extended to other methods that exhibit such properties. 
In the future, we will explore other clustering algorithms that can be used in our exhaustive clustering framework.\vspace{-3mm}

\ifCLASSOPTIONcompsoc
  \section*{Acknowledgments}
\else
  \section*{Acknowledgment}
\fi

We thank our collaborators at Johns Hopkins University, Dr. Ralph Hruban and Dr. Elliot Fishman, for providing us the Abdominal CT and Pancreas datasets. The Chest CT is courtesy Department of Radiology, University of Iowa. The Tomato, Tooth, Vismale, and Test Spheres datasets are from the Volume Dataset Repository maintained by Stephan Rottger at the University of Erlangen-Nuremberg, Germany. This work has been partially supported by the National Science Foundation grants IIS1527200, NRT1633299, CNS1650499, the Marcus Foundation, and the National Heart, Lung, and Blood Institute of the National Institutes of Health under Award Number U01HL127522. The content is solely the responsibility of the authors and does not necessarily represent the official view of the National Institutes of Health. Additional support was provided by the Center for Biotechnology, a New York State Center for Advanced Technology; Cold Spring Harbor Laboratory; Brookhaven National Laboratory; the Feinstein Institute for Medical Research; and the New York State Department of Economic Development under contract C14051.


\vspace{-15mm}
\begin{IEEEbiography}[{\includegraphics[width=1in,height=1.25in,clip,keepaspectratio]{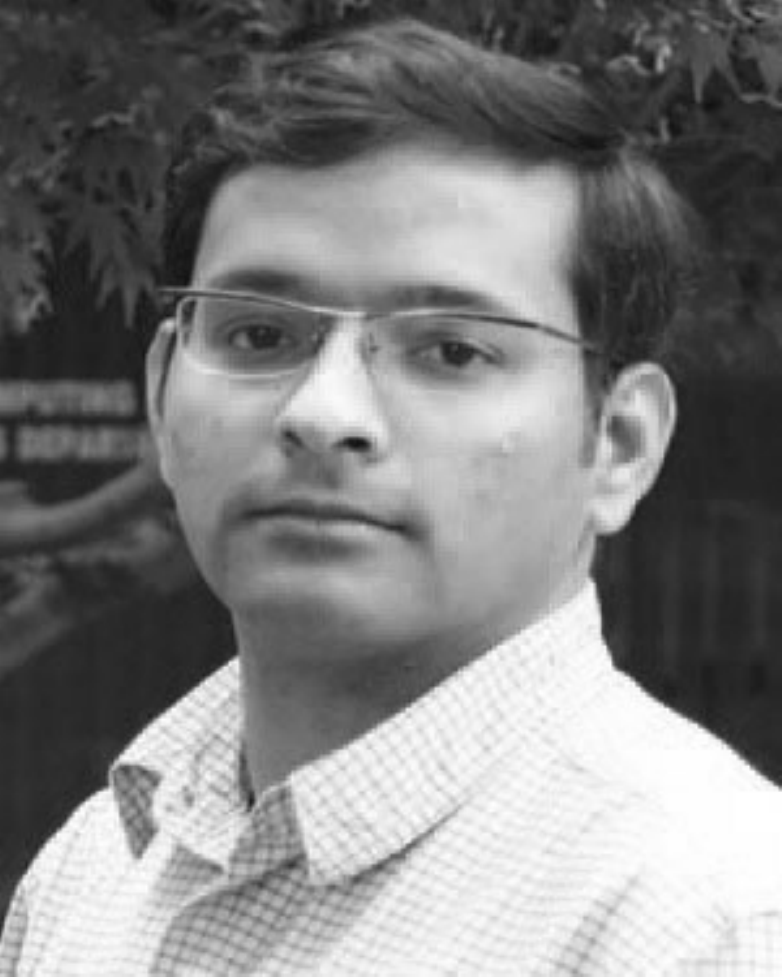}}]{Shreeraj Jadhav}
is a PhD candidate in the Department of Computer Science at Stony Brook University. He did his Bachelor of Mechanical Engineering from the University of Pune, India and his MS in Computing (Graphics and Visualization) from the School of Computing, University of Utah. His research interests include computer graphics and visualization.
\end{IEEEbiography}
\vspace{-15mm}
\begin{IEEEbiography}[{\includegraphics[width=1in,height=1.25in,clip,keepaspectratio]{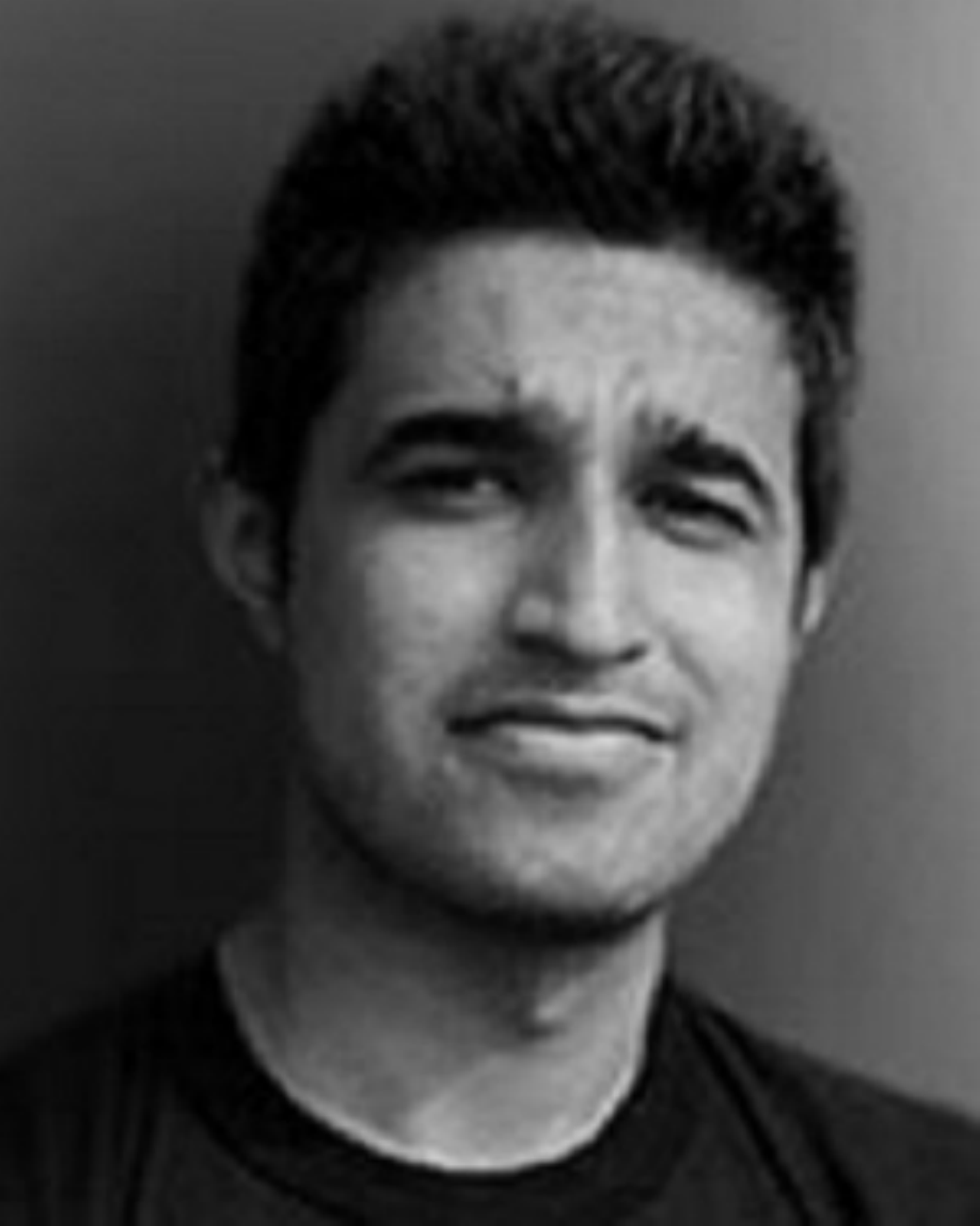}}]{Saad Nadeem}
is a Research Scholar in Department of Medical Physics, Memorial Sloan Kettering Cancer Center. He received his PhD in Computer Science from Stony Brook University, US, in 2017. His research interests include medical imaging, bioinformatics, computer vision, computer graphics, and visualization.
\end{IEEEbiography}
\vspace{-15mm}
\begin{IEEEbiography}[{\includegraphics[width=1in,height=1.25in,clip,keepaspectratio]{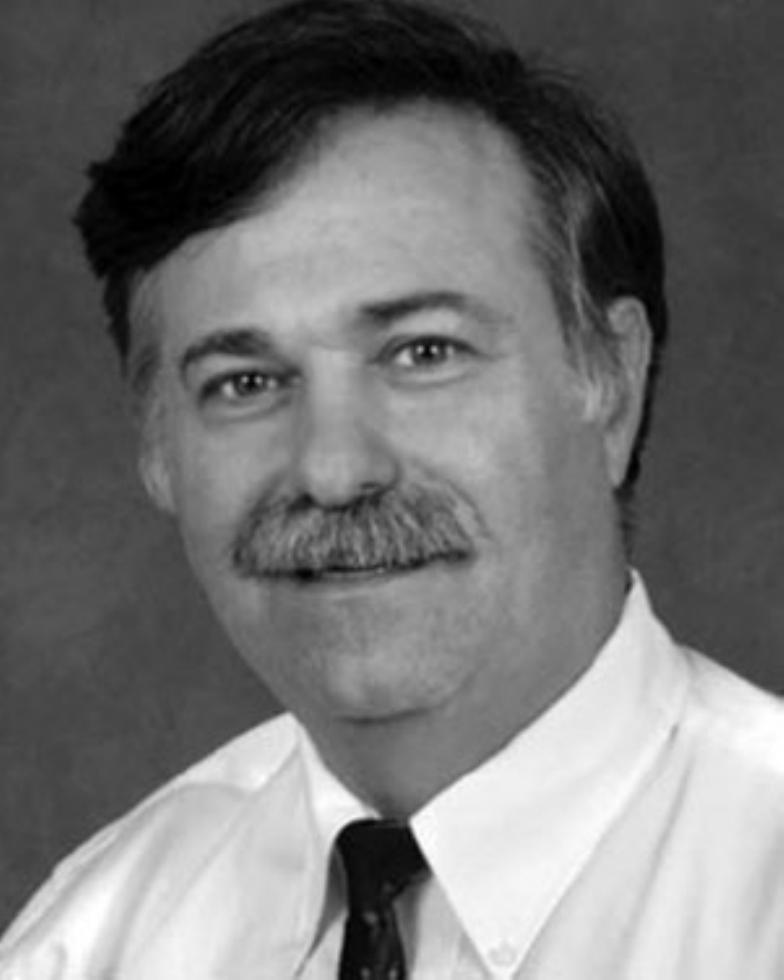}}]{Arie E. Kaufman}
is a Distinguished Professor of Computer Science, Director of the Center of Visual Computing (CVC), and Chief Scientist of the Center of Excellence in Wireless and Information Technology (CEWIT) at Stony Brook University.
He served as Chair of the Computer Science Department 1999-2017.
He has conducted research for over 40 years in visualization and graphics and their applications, has published more than 300 refereed papers, has presented more than 20 invited keynote talks, has been awarded/filed more than 40 patents, and has been PI/co-PI on more than 100 grants. He was the founding Editor-in-Chief of IEEE Transaction on Visualization and Computer Graphics (TVCG), 1995-1998. He is a Fellow of IEEE, a Fellow of ACM, a Fellow of the National Academy of Inventors, the recipient of the IEEE Visualization Career Award (2005), and was inducted into the Long Island Technology Hall of Fame (2013). He received his PhD in Computer Science from Ben-Gurion University, Israel, in 1977.
\end{IEEEbiography}

\end{document}